%% file: iclr2027_conference.tex
\documentclass{article} 
\usepackage{iclr2027_conference,times}

\input{math_commands.tex}

\usepackage{hyperref}
\usepackage{url}
\usepackage{graphicx}   
\usepackage{multirow}   
\usepackage{xcolor}     
\usepackage{colortbl}   
\usepackage{wrapfig}    
\usepackage{float}
\usepackage{booktabs} 
\definecolor{PVYellow}{RGB}{233,242,255} 
\definecolor{PVGreen}{RGB}{0,128,70}    

\newcommand{\pv}[2]{\textbf{#1}\,\textsuperscript{\textcolor{PVGreen}{\ensuremath{\uparrow}#2}}}
\newcommand{\pvn}[2]{#1\,\textsuperscript{\textcolor{PVGreen}{\ensuremath{\uparrow}#2}}}

\title{$\Delta$Representation: Geometry Supervised Representation Learning of Phenotypes via Counterfactual Reasoning for Medical VLMs}

\author{%
\parbox[t]{\dimexpr\textwidth-2\tabcolsep\relax}{%
\centering
\normalfont
\setlength{\tabcolsep}{16pt}
\begin{tabular}[t]{@{}ccc@{}}
\textbf{Hao Wang}$^{1,2}$\thanks{These authors contributed equally to this work and share first authorship.} & \textbf{Qiwei Zeng}$^{2}$\footnotemark[1] & \textbf{Jinghao Lin}$^{3}$ \\[4pt]
\textbf{Shuchang Ye}$^{1}$ & \textbf{Yuezhe Yang}$^{2}$ & \textbf{Yige Peng}$^{2}$ \\[4pt]
\textbf{Haoyuan Che}$^{4}$ & \textbf{Jinman Kim}$^{1}$\thanks{Corresponding authors: Jinman Kim and Lei Bi.} & \textbf{Lei Bi}$^{2}$\footnotemark[2] \\[7pt]
\multicolumn{3}{c}{$^{1}$The University of Sydney} \\
\multicolumn{3}{c}{$^{2}$Shanghai Jiao Tong University} \\
\multicolumn{3}{c}{$^{3}$Northeastern University} \\
\multicolumn{3}{c}{$^{4}$Jilin University}
\end{tabular}%
}%
}

\iclrfinalcopy 
\begin{document}

\maketitle

\begin{abstract}
Medical vision-language models (VLMs) have shown increasing potential for radiological image interpretation.
Medical VLMs encode radiological images into visual representations that capture both anatomical and phenotypic information for diagnosis. 
Existing approaches improve pathological phenotype representations through semantic-guided representation alignment. 
However, pathological phenotypes arise as lesion-specific visual changes superimposed on underlying normal anatomy. 
Such semantic alignment approaches fail to model the phenotype-specific increment relative to the corresponding normal anatomical representation.
To address this gap, we propose \textbf{$\Delta$Representation}, a visual phenotype representation learning framework based on counterfactual reasoning for medical VLMs. 
It comprises \textbf{BaseAnatomy}, a geometry-supervised representation learning module, and \textbf{$\Delta$Phenotype}, a counterfactual incremental representation learning module. 
BaseAnatomy provides fine-grained geometric supervision through spatial relationships across and within anatomical structures. 
$\Delta$Phenotype computes the representation increment between lesion representations and their corresponding normal anatomical representations, and supervises increments associated with the same phenotype to cluster in the representation space.
Experiments on \textit{ReXGroundingCT} and \textit{LIDC-IDRI} demonstrate that $\Delta$Representation effectively structures pathological phenotype representations and improves lesion grounding and phenotype characterization accuracy in medical VLMs. 
Code is available at https://anonymous.4open.science/r/deltarep-CF6D.
\end{abstract}

\section{Introduction}

Medical vision-language models (VLMs) have been developed to assist clinicians in diagnosis based on radiological images \cite{wu2025towards,xu2026lingshu}. 
Through large-scale image-text pretraining, medical VLMs establish associations between radiological images and clinical texts using vision and text encoders. 
The vision encoder extracts pathology-related visual information into representations, which are subsequently aligned with semantic representations to support diagnostic reasoning.
However, vision encoders trained under this paradigm lack explicit modeling of anatomical structures and pathological visual phenotypes in the visual representation space.
Consequently, potentially inaccurate visual representations can be aligned with semantic representations and used as evidence for diagnosis.

To reduce such incorrect associations, existing studies have explored fine-grained representation alignment to connect visual phenotypic representations with clinical semantic representations. 
TRRG \cite{wang2025trrg} uses report sentences as semantic supervision and aligns sentence embeddings with image-level visual features through contrastive learning. 
PRIOR \cite{cheng2023prior} further performs local alignment between regional visual features and report sentences through attention mechanisms, establishing finer-grained correspondences between abnormalities and their textual descriptions. 
MedKLIP \cite{wu2023medklip} introduces pathology concepts extracted from radiology reports and uses them to guide the learning of disease-specific visual representations. 
MCA-RG \cite{xing2025mca} aligns visual features with pathology concepts and introduces pathology--anatomy matching to emphasize disease-related regions. 
RadAlign \cite{gu2025radalign} uses learnable concept tokens to extract attribute-specific visual features and aligns them with textual descriptions of diagnostic criteria through contrastive learning.

However, learning from paired image-text data cannot explicitly establish the incremental relationship between pathological phenotypes and their underlying anatomical structures. 
Semantic representations derived from text describe the presence and subclasses of pathological phenotypes. They do not explicitly represent the phenotypic changes relative to the corresponding normal anatomy. 
Moreover, these semantic representations provide limited supervision for separating visually similar phenotypes in the representation space. 
This limits the vision encoder’s ability to learn visual changes associated with pathological phenotypes and distinguish different phenotypes. 
To establish this relationship directly in the visual representation space, we define a pathological phenotype as the visual change relative to its underlying normal anatomy and construct this phenotype representation by comparing a pathological patch with its corresponding normal counterpart.

\begin{figure}
    \centering
    \includegraphics[width=\linewidth]{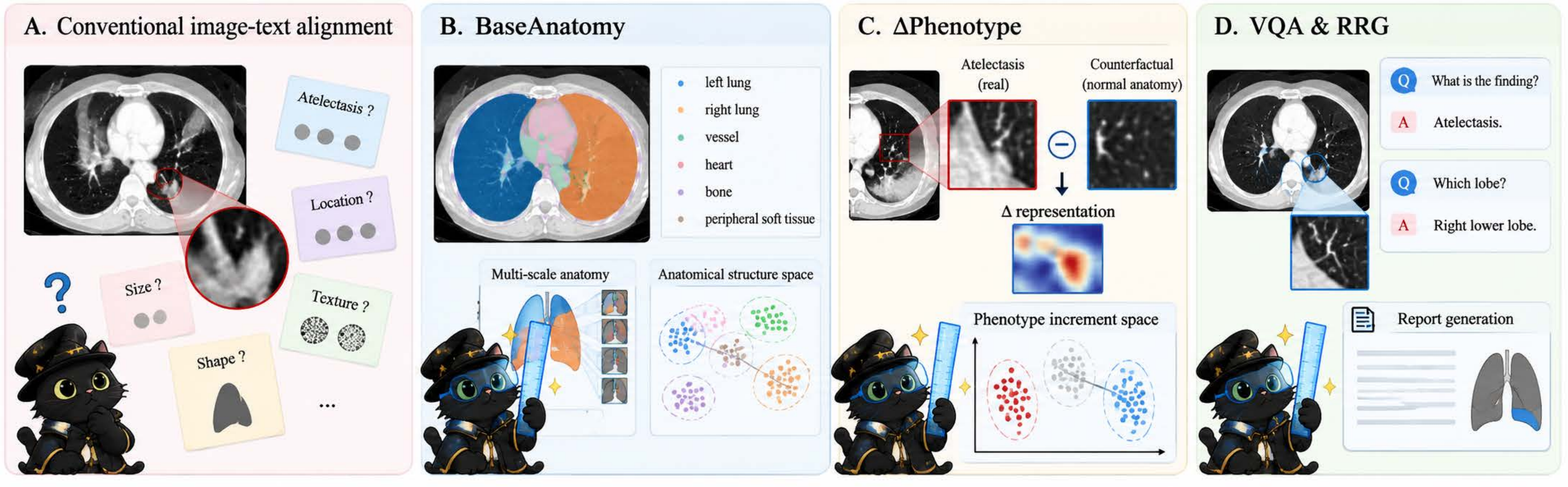}
    \vspace{-15pt}
    \caption{BaseAnatomy structures the anatomical visual representation space, while $\Delta$Phenotype organizes lesion-specific increments relative to counterfactual normal anatomy.}
    \label{fig:overview}
    \vspace{-15pt}
\end{figure}

To implement this theory, we propose \textbf{$\Delta$Representation}, a visual representation learning framework for medical VLMs that integrates anatomical geometry supervision with counterfactual phenotype learning as shown in Figure~\ref{fig:overview}. 
First, \textbf{BaseAnatomy} supervises visual embeddings using ideal spatial distributions that preserve geometric relationships between anatomical structures and spatial continuity within each structure. 
This establishes stable normal anatomical representations as references for learning pathological changes. 
Building on these references, \textbf{$\Delta$Phenotype} uses counterfactual reasoning to learn phenotype-specific increments between the representations of pathological patches and their corresponding normal counterparts. 
These increments explicitly characterize pathological visual changes relative to normal anatomy. 
Experiments on \textit{ReXGroundingCT} and \textit{LIDC-IDRI} demonstrate that $\Delta$Representation consistently improves lesion grounding and phenotype characterization across medical VLMs. 
Representation analyses further show that the learned phenotype increments provide more discriminative visual representations for different pathological phenotypes. 
Our contributions are summarized as follows:

\noindent \textbf{(1)} We propose \textbf{$\Delta$Representation}, a visual representation learning framework for medical VLMs that combines anatomical geometry supervision with counterfactual phenotype learning. It models pathological phenotypes as representation increments relative to normal anatomy.

\noindent \textbf{(2)} We develop \textbf{BaseAnatomy}, which organizes normal anatomical representations using ideal spatial distributions. It preserves inter-structure geometry and intra-structure spatial continuity to provide stable anatomical references.

\noindent \textbf{(3)} We introduce \textbf{$\Delta$Phenotype}, which learns phenotype-specific increments through counterfactual reasoning. It progressively refines phenotype clusters as additional samples become available, improving intra-phenotype consistency and inter-phenotype separation.

\section{Related Work}

\subsection{Medical VLMs in Radiology}

Medical VLMs adapt pretrained visual and language models to medical interpretation through domain-specific multimodal training. LLaVA-Med combines biomedical figure-caption alignment with instruction tuning to support medical visual conversation \cite{li2023llava}. LLaVA-Rad connects pretrained image and language components through a lightweight adapter and specializes the model using radiological image--text pairs \cite{zambrano2025clinically}. MedGemma incorporates medical image encoder adaptation into a Gemma-based multimodal model \cite{sellergren2025medgemma}. Complementing model adaptation, MedRAX integrates chest X-ray analysis tools with multimodal models to address complex clinical queries \cite{fallahpour2025medrax}.

The scope of radiology VLMs also extends to volumetric inputs and anatomically structured interpretation. M3D-LaMed unifies multiple 3D tasks, including report generation, question answering, positioning, and segmentation, within a multimodal instruction-learning framework \cite{bai2024m3d}. Argus studies how encoder pretraining, visual token compression, and model and data scaling affect high-resolution CT report generation \cite{liu2025argus}. AOR introduces anatomy-centric, region-level reasoning to structure chest X-ray interpretation \cite{li2026aor}.

\subsection{Fine-Grained Alignment in Medical VLMs}

Fine-grained alignment methods link local image regions to clinically meaningful descriptions. BiRD learns region--language correspondence from grounding instructions constructed from segmentation datasets, while MAIRA-2 associates radiological findings with bounding boxes during report generation \cite{huang2024refer,bannur2024maira}. Anatomy-aware methods further organize visual inputs around explicit anatomical regions. Reg2RG uses segmentation masks to extract regional features for region-specific report generation, whereas MedRegion-CT combines region-guided tokenization with pseudo-mask supervision and structured lesion attributes \cite{chen2025large,kyung2025medregion}. These approaches improve local correspondence through regional feature construction, explicit supervision, or region-specific inputs to the language model.

Concept-guided methods further associate visual representations with clinically meaningful semantics. MCA-RG introduces anatomical and pathological concept banks, extracts concept-specific visual features through cross-attention, and combines contrastive learning with feature gating \cite{xing2025mca}. RadAlign aligns visual concept tokens with textual diagnostic criteria to support disease classification and report generation \cite{gu2025radalign}. Although these methods strengthen regional and concept-specific correspondence, alignment with text-derived anatomical semantics does not explicitly constrain visual embeddings to preserve geometric relationships between anatomical structures or spatial continuity within each structure. Furthermore, phenotypic semantic supervision does not explicitly characterize the counterfactual changes between pathological patches and their normal counterparts, leaving phenotype-specific representation increments unmodeled. These limitations hinder stable anatomical encoding and the learning of phenotype-specific changes relative to normal anatomy, limiting the vision encoder's ability to accurately represent pathological visual phenotypes.

\section{Methodology}

\subsection{Overview}

\begin{figure}
    \centering
    \includegraphics[width=\linewidth]{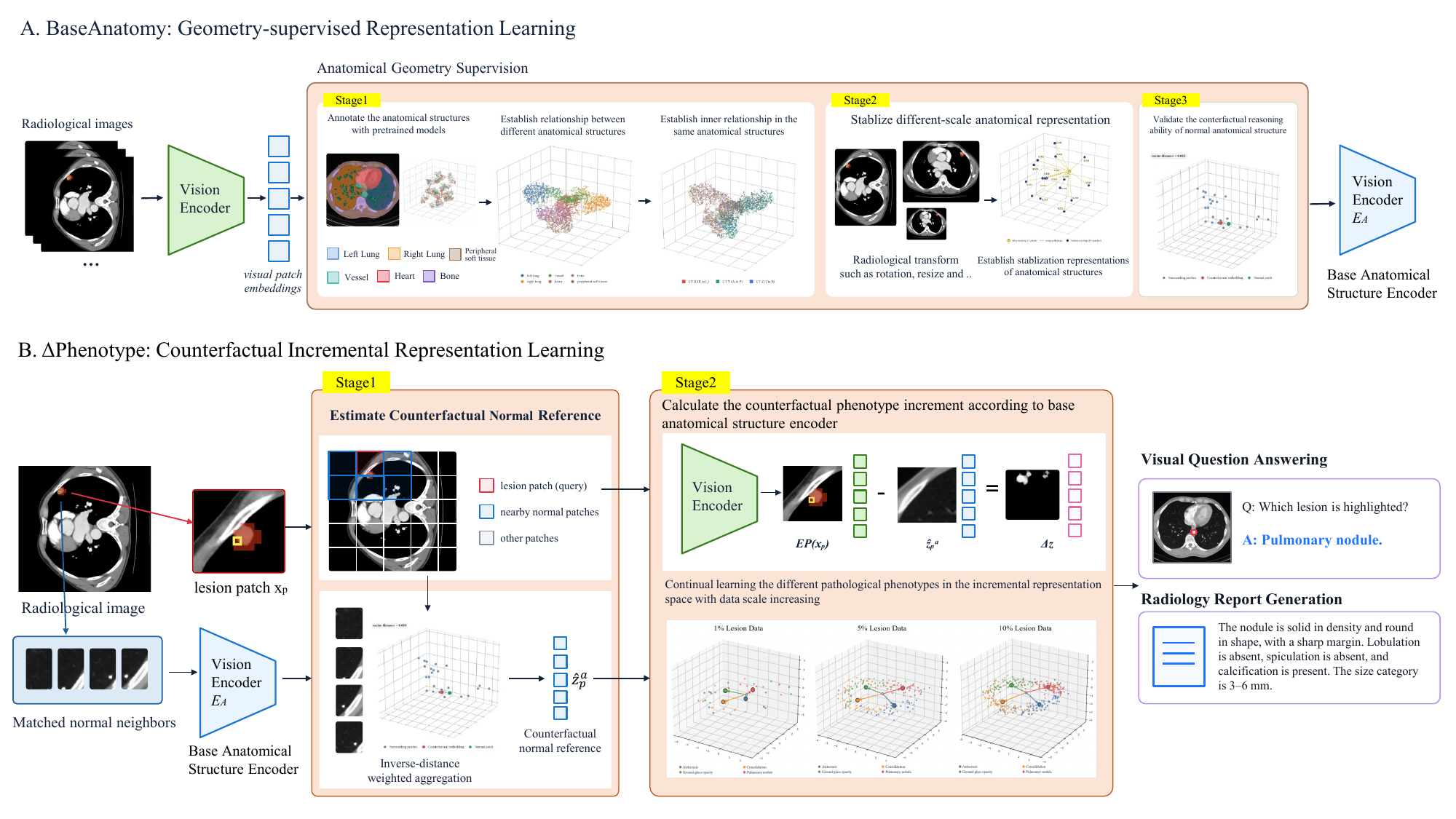}
    \vspace{-15pt}
    \caption{Overview of $\Delta$Representation. (a) BaseAnatomy shapes an anatomically structured visual space with ideal geometric supervision that encodes anatomical relationships and spatial continuity. (b) $\Delta$Phenotype constructs counterfactual normal references for lesion patches and learns phenotype-specific visual increments relative to normal anatomy for discriminative phenotype representation learning.}
    \label{fig:flow}
    \vspace{-15pt}
\end{figure}

$\Delta$Representation consists of two components as shown in Figure~\ref{fig:flow}. \textbf{BaseAnatomy} shapes the normal anatomical representation space using ideal spatial distributions, which preserve geometric relationships between anatomical structures and spatial continuity within each structure. \textbf{$\Delta$Phenotype} then constructs counterfactual normal references for pathological patches and learns phenotype-specific representation increments by comparing pathological representations with their corresponding normal representations. These increments are progressively refined through phenotype-specific clustering as additional lesion samples become available. BaseAnatomy is trained first, after which its anatomical encoder is frozen to provide stable normal references during $\Delta$Phenotype learning.

\subsection{\textbf{BaseAnatomy}: Building Geometry-Supervised Normal Anatomical Representation Space}

BaseAnatomy learns an anatomically organized representation space with a visual encoder $E_A$, using annotation-defined ideal spatial distributions to preserve both relationships across anatomical structures and spatial continuity within each structure. Given a CT image $\mathbf{x}$, $E_A$ extracts $\ell_2$-normalized patch representations $\{\mathbf{a}_p\}_{p=1}^{P}$. Each normal patch is associated with an anatomical label $c_p$ and normalized CT coordinates $\mathbf{u}_p$, while slice-position encoding provides its relative position along the volumetric axis. Anatomical annotations define tissue regions, and lesion masks exclude pathological patches from normal anatomical supervision.

\paragraph{Ideal anatomical geometry supervision.}
We construct the ideal representation space from two complementary spatial relationships. Inter-structure geometry preserves the relative positions between different anatomical structures, while intra-structure geometry preserves the spatial arrangement of patches within the same structure. Specifically, the target distance between normal patches $p$ and $q$ is defined as
\begin{equation}
    d^{\star}_{pq}
    =
    \sqrt{
    \alpha\,d_{A}(c_p,c_q)^2
    +(1-\alpha)\lVert\mathbf{u}_p-\mathbf{u}_q\rVert_2^2
    },
    \label{eq:ideal_anatomical_geometry}
\end{equation}
where $d_A(c_p,c_q)$ denotes the normalized distance between the centers of their anatomical structures, and $\alpha$ balances inter-structure geometry and intra-structure spatial continuity. For patches from the same anatomical structure, $d_A(c_p,c_q)=0$, such that their target relationship is determined by their relative spatial positions. This ideal distribution provides direct geometric supervision for anatomical representations without relying on text-derived anatomical semantics.

\paragraph{Geometry-supervised anatomical representation learning.}
We optimize $E_A$ by matching pairwise cosine distances between patch representations to their corresponding ideal distances using a smooth-$L_1$ objective. Anatomical compactness encourages patches belonging to the same structure to form coherent distributions, while neighborhood constraints preserve local spatial continuity within each structure. We further apply rotation and scale transformations with matched spatial correspondence, encouraging representations of the same anatomical region to remain consistent across transformed views. Anatomical classes are sampled in a balanced manner to prevent large structures from dominating the learned space. Together, these constraints establish stable normal anatomical representations that serve as references for subsequent phenotype learning. Anatomical annotations and lesion masks are used only as supervision during training.

\subsection{\textbf{$\Delta$Phenotype}: Learning Counterfactual Phenotype Increments from Normal Anatomy}

$\Delta$Phenotype learns pathological visual phenotypes as representation increments relative to their underlying normal anatomy. We initialize a frozen anatomical encoder $E_A$ and a trainable phenotype encoder $E_P$ from the same BaseAnatomy checkpoint. $E_A$ preserves the learned anatomical representation space and provides stable normal references, while $E_P$ is optimized to encode phenotype-specific pathological changes. Only the final vision layers and normalization layer of $E_P$ are updated during phenotype learning.After representation learning, only $E_P$ is retained as the VLM vision encoder for downstream inference; $E_A$ and all anatomical/lesion annotations are used only during training.

\paragraph{Counterfactual normal reference and phenotype increment.}
For each pathological patch, we construct a counterfactual representation of the normal anatomy at the same location. Since BaseAnatomy preserves spatial continuity within anatomical structures, this normal counterpart can be estimated from nearby mask-negative patches belonging to the corresponding anatomical region. For a pathological patch $p$, we define
\begin{equation}
    \hat{\mathbf{a}}_p
    =
    \operatorname{norm}\!\left(
    \sum_{q\in\mathcal{N}(p)}
    w_{pq}\mathbf{a}_q
    \right),
    \qquad
    \boldsymbol{\delta}_p
    =
    E_P(\mathbf{x})_p-\hat{\mathbf{a}}_p,
    \label{eq:counterfactual_phenotype_increment}
\end{equation}
where $\mathcal{N}(p)$ contains anatomically matched normal neighbors, $w_{pq}$ denotes normalized inverse-distance weights, and $\hat{\mathbf{a}}_p$ represents the estimated normal anatomical representation. The resulting increment $\boldsymbol{\delta}_p$ isolates the pathological visual change relative to this counterfactual normal reference.

\paragraph{Phenotype-specific clustering and continual refinement.}
We organize the learned increments according to their pathological phenotype labels. Patch-level increments from the same annotated lesion are aggregated into a region-level phenotype representation. Increments associated with the same phenotype are encouraged to remain consistent, while those from different phenotypes are separated in the representation space. Localization supervision further distinguishes lesion increments from hard normal patches, while anatomical preservation constrains normal-patch representations of $E_P$ to remain close to those produced by the frozen $E_A$. No-lesion samples additionally suppress pathological responses in normal tissues.

Starting from limited lesion samples, these objectives establish an initial organization of phenotype-specific increments. As additional lesion samples become available, their increments are incorporated into the same learning process, continually refining the phenotype distributions while the normal anatomical reference space remains fixed. This enables $E_P$ to progressively improve the encoding and discrimination of pathological visual changes across different anatomical contexts.
\section{Experiments}

\begin{table*}[!t]
  \centering
  \tiny
  \setlength{\tabcolsep}{8pt}
  \renewcommand{\arraystretch}{1.18}
  \resizebox{\textwidth}{!}{
  \begin{tabular}{@{\hspace{4pt}}l|c|cc|cc@{\hspace{4pt}}}
    \hline\hline
    \multirow{2}{*}{\textbf{Models}} & \multirow{2}{*}{\textbf{Scale}} & \multicolumn{2}{c|}{\textbf{ReXGroundingCT}} & \multicolumn{2}{c}{\textbf{LIDC-IDRI}} \\ \cline{3-6}
    & & Grounding Acc. & Phenotype Acc. & Grounding Acc. & Phenotype Acc. \\ \hline
    RadFM & 14B & 10.60 & 4.95 & 18.18 & 2.48 \\
    \rowcolor{PVYellow} \quad \textbf{w/ $\Delta$Representation} & $\sim$14B & \pv{50.53}{39.93} & \pv{38.87}{33.92} & \pv{62.81}{44.63} & \pv{47.11}{44.63} \\ \hline
    LLaVA-Med & 7B & 43.11 & 12.37 & 2.48 & 5.79 \\
    \rowcolor{PVYellow} \quad \textbf{w/ $\Delta$Representation} & $\sim$7B & \pv{66.78}{23.67} & \pv{51.94}{39.57} & \pv{47.11}{44.63} & \pv{44.63}{38.84} \\ \hline
    Lingshu & 7B & 77.39 & 15.19 & 23.97 & 42.15 \\
    \rowcolor{PVYellow} \quad \textbf{w/ $\Delta$Representation} & $\sim$7B & \pv{91.17}{13.78} & \pv{35.69}{20.50} & \pv{65.29}{41.32} & \pv{86.78}{44.63} \\ \hline
    Hulu-Med & 4B & 70.32 & 26.15 & 24.79 & 41.32 \\
    \rowcolor{PVYellow} \quad \textbf{w/ $\Delta$Representation} & $\sim$4B & \pv{86.93}{16.61} & \pv{39.22}{13.07} & \pv{60.33}{35.54} & \pv{85.95}{44.63} \\ \hline
    MedGemma & 4B & 65.37 & 16.96 & 52.89 & 42.98 \\
    \rowcolor{PVYellow} \quad \textbf{w/ $\Delta$Representation} & $\sim$4B & \pv{79.15}{13.78} & \pv{35.69}{18.73} & \pv{73.55}{20.66} & \pv{85.12}{42.14} \\ \hline
    MedGemma1.5 & 4B & 43.82 & 15.55 & 33.06 & 43.80 \\
    \quad w/ RadAlign & $\sim$4B & 46.29 & 14.84 & 54.55 & 34.71 \\
    \quad w/ MCA-RG & $\sim$4B & 46.29 & 40.64 & 54.55 & 36.36 \\
    \quad w/ Reg2RG & $\sim$4B & 45.94 & 40.64 & 47.93 & 36.36 \\
    \rowcolor{PVYellow} \quad \textbf{w/ $\Delta$Representation} & $\sim$4B & \pv{79.86}{36.04} & \pv{43.11}{27.56} & \pv{76.86}{43.80} & \pv{77.69}{33.89} \\ \hline\hline
  \end{tabular}}
  \caption{Comparison with pretrained medical VLMs and alignment methods on the VQA task.}
  \label{tab:vqa_comparison}

  \vspace{8pt}

  \resizebox{\textwidth}{!}{
  \begin{tabular}{@{\hspace{4pt}}l|c|cc|cc@{\hspace{4pt}}}
    \hline\hline
    \multirow{2}{*}{\textbf{Models}} & \multirow{2}{*}{\textbf{Scale}} & \multicolumn{2}{c|}{\textbf{ReXGroundingCT}} & \multicolumn{2}{c}{\textbf{LIDC-IDRI}} \\ \cline{3-6}
    & & Grounding Acc. & Phenotype Acc. & Grounding Acc. & Phenotype Acc. \\ \hline
    RadFM & 14B & 3.18 & 4.95 & 4.96 & 6.61 \\
    \rowcolor{PVYellow} \quad \textbf{w/ $\Delta$Representation} & $\sim$14B & \pv{14.13}{10.95} & \pv{15.55}{10.60} & \pv{34.71}{29.75} & \pv{37.19}{30.58} \\ \hline
    LLaVA-Med & 7B & 4.95 & 41.34 & 14.05 & 58.68 \\
    \rowcolor{PVYellow} \quad \textbf{w/ $\Delta$Representation} & $\sim$7B & \pv{19.79}{14.84} & \pv{57.24}{15.90} & \pv{52.07}{38.02} & \pv{94.21}{35.53} \\ \hline
    Lingshu & 7B & 0.71 & 3.89 & 22.31 & 35.54 \\
    \rowcolor{PVYellow} \quad \textbf{w/ $\Delta$Representation} & $\sim$7B & \pv{12.01}{11.30} & \pv{20.49}{16.60} & \pv{48.76}{26.45} & \pv{74.38}{38.84} \\ \hline
    Hulu-Med & 4B & 2.47 & 8.48 & 38.84 & 42.15 \\
    \rowcolor{PVYellow} \quad \textbf{w/ $\Delta$Representation} & $\sim$4B & \pv{12.72}{10.25} & \pv{18.73}{10.25} & \pv{52.07}{13.23} & \pv{86.78}{44.63} \\ \hline
    MedGemma & 4B & 0.35 & 3.53 & 13.22 & 22.31 \\
    \rowcolor{PVYellow} \quad \textbf{w/ $\Delta$Representation} & $\sim$4B & \pv{14.49}{14.14} & \pv{15.90}{12.37} & \pv{44.63}{31.41} & \pv{66.94}{44.63} \\ \hline
    MedGemma1.5 & 4B & 0.71 & 5.30 & 29.75 & 39.67 \\
    \quad w/ RadAlign & $\sim$4B & 18.02 & 43.46 & 28.93 & 30.58 \\
    \quad w/ MCA-RG & $\sim$4B & 26.15 & 51.24 & 34.71 & 38.02 \\
    \quad w/ Reg2RG & $\sim$4B & 26.15 & 51.24 & 34.71 & 38.02 \\
    \rowcolor{PVYellow} \quad \textbf{w/ $\Delta$Representation} & $\sim$4B & \pv{29.33}{28.62} & \pv{52.65}{47.35} & \pv{50.41}{20.66} & \pv{75.21}{35.54} \\ \hline\hline
  \end{tabular}}
  \caption{Comparison with pretrained medical VLMs and alignment methods on the RRG task.}
  \label{tab:rrg_comparison}
  \vspace{-15pt}
\end{table*}

\subsection{Datasets and Evaluation}
We conduct training and evaluation on two public chest CT datasets, \textbf{ReXGroundingCT} \cite{baharoon2026rexgroundingct} and \textbf{LIDC-IDRI} \cite{armato2011lung}, which provide complementary annotations for pulmonary abnormalities. From ReXGroundingCT, we select 2,720 training, 230 development, and 283 test axial CT images covering four target phenotypes: atelectasis, consolidation, ground-glass opacity, and pulmonary nodule. From LIDC-IDRI, we use 681 training, 164 development, and 121 test images; the test set includes 71 nodule-positive and 50 no-lesion cases. All splits are patient-disjoint. For each image, we construct two visual question answering (VQA) questions and one radiology report generation (RRG) prompt. The test set therefore comprises 808 VQA questions (566 from ReXGroundingCT and 242 from LIDC-IDRI) and 404 RRG instances (283 and 121, respectively). For lesion localization, we divide each image into a $4 \times 4$ grid and define the reference cell as the one containing the largest annotated lesion area; no-lesion cases have a dedicated \textit{No lesion} answer. We report \textit{Grounding Accuracy} and \textit{Phenotype Accuracy} for both VQA and RRG. Grounding Accuracy measures exact 4×4 grid-cell localization for VQA and the correctness of lesion-location attributes for RRG, while Phenotype Accuracy measures phenotype identification.Appendix~\ref{app:data}.

\subsection{Experimental Setup}
\begin{figure}
    \centering
    \includegraphics[width=\linewidth]{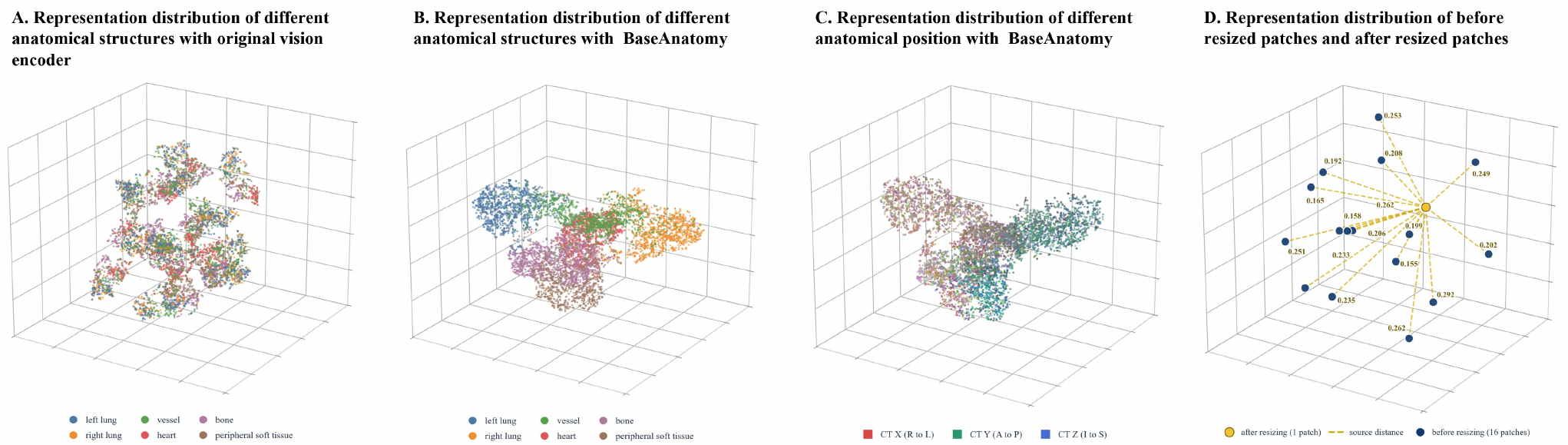}
    \vspace{-10pt}
    \caption{Anatomical representation analysis of BaseAnatomy. (A–B) Representation distributions before and after BaseAnatomy training. (C) Spatial continuity across anatomical positions. (D) Cross-scale consistency of anatomical representations.}
    \label{fig:anatomical_representation_space}
    \vspace{-10pt}
\end{figure}

\begin{figure}
    \centering
    \includegraphics[width=\linewidth]{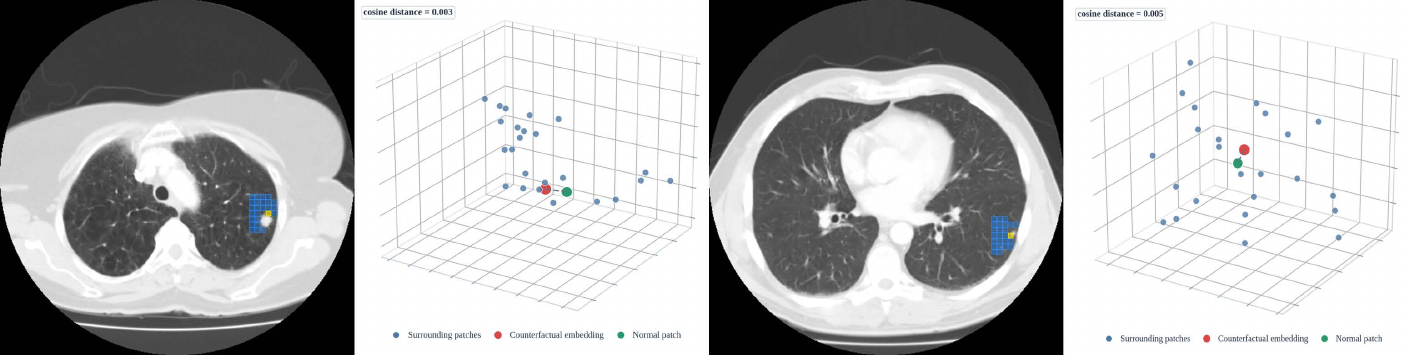}
    \vspace{-10pt}
    \caption{Counterfactual representation analysis. Comparison between inferred counterfactual representations and their corresponding normal representations.}
    \label{fig:counterfactual_reasoning}
    \vspace{-15pt}
\end{figure}

\begin{figure}
    \centering
    \includegraphics[width=\linewidth]{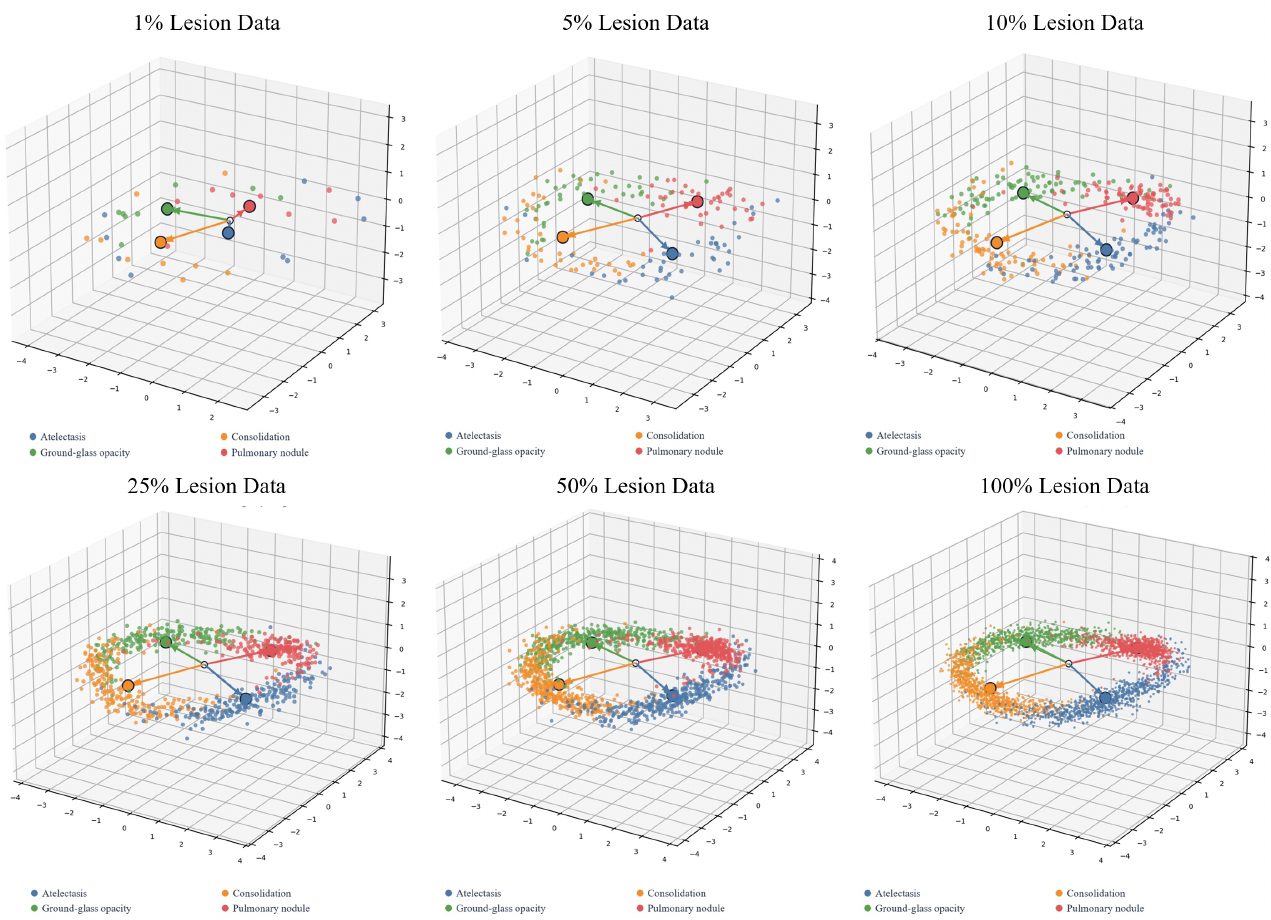}
    \vspace{-10pt}
    \caption{Representation learning performance of visual phenotypes in different lesion data scales.}
    \label{fig:different_scale_performance}
    \vspace{-15pt}
\end{figure}

\begin{figure}
    \centering
    \includegraphics[width=\linewidth]{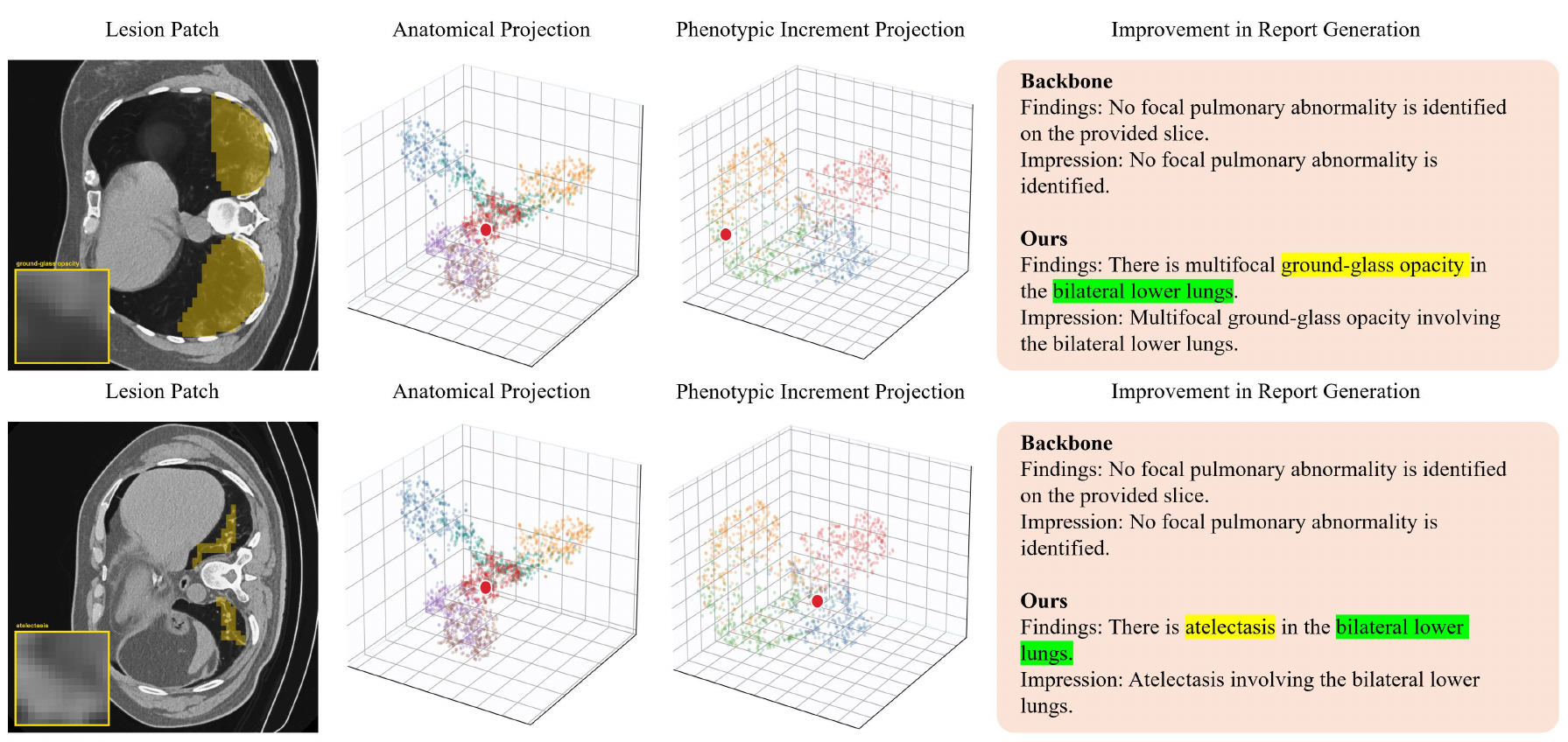}
    \vspace{-15pt}
    \caption{Case study of $\Delta$Representation, illustrating anatomical projection by BaseAnatomy, phenotype-increment projection by $\Delta$Phenotype, and their contribution to report generation.}
    \label{fig:case_study}
    \vspace{-15pt}
\end{figure}

We conduct a comprehensive evaluation of $\Delta$Representation from both downstream performance and module effectiveness. First, we compare $\Delta$Representation across six pretrained medical VLMs \cite{wu2025towards,li2023llava,xu2026lingshu,jiang2025hulu,sellergren2025medgemma,sellergren2026medgemma} and with existing representation alignment methods \cite{gu2025radalign,xing2025mca,chen2025large} on VQA and RRG tasks to evaluate anatomical grounding and pathological phenotype characterization (Tables~\ref{tab:vqa_comparison} and~\ref{tab:rrg_comparison}). For each VLM, $\Delta$Representation is initialized from and trained on its
native vision encoder, while retaining the corresponding language backbone. We then analyze BaseAnatomy by visualizing anatomical representations before and after training, examining their spatial continuity and cross-scale consistency, and comparing inferred counterfactual normal representations with their corresponding real normal representations (Figs.~\ref{fig:anatomical_representation_space} and~\ref{fig:counterfactual_reasoning}). For $\Delta$Phenotype, we progressively increase lesion-positive training data from 1\% to 100\% and evaluate both downstream performance and phenotype representation distributions, assessing whether additional samples continually refine intra-phenotype consistency and inter-phenotype separation (Table~\ref{tab:lesion_data_ratio} and Fig.~\ref{fig:different_scale_performance}). Finally, we provide an end-to-end case study to illustrate how anatomical representations and phenotype-specific increments support pathological characterization and report generation (Fig.~\ref{fig:case_study}).

\subsection{Comparison with Different Medical VLMs and Alignment Methods}

Table~\ref{tab:vqa_comparison} presents the comparison on the VQA task. To evaluate the effectiveness and generalizability of $\Delta$Representation, we apply it to six pretrained medical VLM backbones on ReXGroundingCT and LIDC-IDRI. $\Delta$Representation consistently improves anatomical grounding and pathological phenotype characterization across all evaluated backbones and datasets. For example, with MedGemma1.5, the absolute gains in grounding and phenotype accuracies are 36.04\% and 27.56\% on ReXGroundingCT, and 43.80\% and 33.89\% on LIDC-IDRI, respectively. Averaged across the six backbones and two datasets, $\Delta$Representation achieves absolute improvements of 31.20\% in grounding accuracy and 33.51\% in phenotype accuracy, demonstrating its effectiveness across different pretrained medical VLMs. Furthermore, we compare $\Delta$Representation with representative visual--text alignment methods, including RadAlign, MCA-RG, and Reg2RG, under the same MedGemma1.5 backbone. $\Delta$Representation achieves the highest grounding and phenotype accuracies on both datasets among these methods. The proposed framework combines anatomical supervision based on ideal spatial distributions with counterfactual learning of phenotype-specific representation increments. These results support its effectiveness in improving anatomical localization and pathological phenotype characterization.

Table~\ref{tab:rrg_comparison} further reports the comparison on the RRG task. To evaluate the effectiveness of $\Delta$Representation for open-ended RRG, we apply it to the same six pretrained medical VLM backbones. $\Delta$Representation consistently improves anatomical grounding and pathological phenotype characterization across both datasets, demonstrating that its benefits extend to free-form report generation. For example, with MedGemma1.5, the absolute gains in grounding and phenotype accuracies are 28.62\% and 47.35\% on ReXGroundingCT, and 20.66\% and 35.54\% on LIDC-IDRI, respectively. Similar improvements are observed across the other backbones, with average absolute gains of 20.80\% in grounding accuracy and 28.57\% in phenotype accuracy across all backbone--dataset combinations. Furthermore, compared with RadAlign, MCA-RG, and Reg2RG under the same MedGemma1.5 backbone, $\Delta$Representation achieves the highest grounding and phenotype accuracies on both datasets. These results demonstrate the effectiveness of the proposed visual representation learning framework in improving the anatomical and pathological accuracy of generated reports. Detailed phenotype evaluation results are provided in Appendix~\ref{app:diseasewise}.

\begin{table*}
  \centering
  \tiny
  \setlength{\tabcolsep}{1pt}
  \renewcommand{\arraystretch}{1.5}
  \resizebox{\textwidth}{!}{
  \begin{tabular}{@{\hspace{4pt}}c|cc|cc|cc|cc@{\hspace{4pt}}}
    \hline\hline
    \multirow{2}{*}{\textbf{Lesion Data Scale}} & \multicolumn{2}{c|}{\textbf{ReXGroundingCT (VQA)}} & \multicolumn{2}{c|}{\textbf{ReXGroundingCT (RRG)}} & \multicolumn{2}{c|}{\textbf{LIDC-IDRI (VQA)}} & \multicolumn{2}{c}{\textbf{LIDC-IDRI (RRG)}} \\ \cline{2-9}
    & Grounding Acc. & Phenotype Acc. & Grounding Acc. & Phenotype Acc. & Grounding Acc. & Phenotype Acc. & Grounding Acc. & Phenotype Acc. \\ \hline
    1\% & 54.06 & 25.80 & 10.95 & 15.55 & 43.80 & 54.55 & 40.50 & 50.41 \\
    5\% & \pvn{72.44}{18.38} & \pvn{34.63}{8.83} & \pvn{10.95}{0.00} & \pvn{15.55}{0.00} & \pvn{44.63}{0.83} & \pvn{62.81}{8.26} & \pvn{40.50}{0.00} & \pvn{51.24}{0.83} \\
    10\% & \pvn{75.27}{2.83} & \pvn{39.93}{5.30} & \pvn{22.26}{11.31} & \pvn{31.45}{15.90} & \pvn{60.33}{15.70} & \pvn{68.60}{5.79} & \pvn{43.80}{3.30} & \pvn{53.72}{2.48} \\
    25\% & \pvn{78.09}{2.82} & \pvn{40.64}{0.71} & \pvn{23.32}{1.06} & \pvn{43.11}{11.66} & \pvn{73.55}{13.22} & \pvn{77.69}{9.09} & \pvn{44.63}{0.83} & \pvn{69.42}{15.70} \\
    50\% & \pvn{79.15}{1.06} & \pvn{43.11}{2.47} & \pvn{28.27}{4.95} & \pvn{50.18}{7.07} & \pvn{76.03}{2.48} & \pvn{77.69}{0.00} & \pvn{47.11}{2.48} & \pvn{75.21}{5.79} \\
    \textbf{100\%} & \pv{79.86}{0.71} & \pv{43.11}{0.00} & \pv{29.33}{1.06} & \pv{52.65}{2.47} & \pv{76.86}{0.83} & \pv{77.69}{0.00} & \pv{50.41}{3.30} & \pv{75.21}{0.00} \\ \hline\hline
  \end{tabular}}
  \caption{Performance on ReXGroundingCT and LIDC-IDRI with different proportions of lesion data for the VQA and RRG tasks.}
  \label{tab:lesion_data_ratio}
  \vspace{-15pt}
\end{table*}

\subsection{Spatial Organization and Scale Consistency of Visual Anatomical Representations with BaseAnatomy}
\label{sec:anatomical_spatial_structure}

Figure~\ref{fig:anatomical_representation_space} compares anatomical patch representations before and after training. With the original vision encoder, patches from different anatomical structures are scattered and substantially intermixed in the t-SNE projection (Fig.~\ref{fig:anatomical_representation_space}A). After training, patches form more coherent clusters whose relative arrangement broadly reflects the spatial relationships among anatomical structures (Fig.~\ref{fig:anatomical_representation_space}B). Within each structure, patch representations also vary gradually with their positions along the three CT spatial axes (Fig.~\ref{fig:anatomical_representation_space}C), indicating that the encoder retains fine-grained location information rather than anatomical identity alone. We further examine whether these representations remain stable across image scales. After downsampling the image, a single patch covers the region occupied by 16 patches at the original scale. Its representation lies near the center of the original patches' representation distribution (Fig.~\ref{fig:anatomical_representation_space}D). This cross-scale consistency suggests that learning fine-grained anatomical spatial relationships enables the vision encoder to produce stable anatomical representations despite changes in image scale. 

We further validate the counterfactual normal representations constructed for pathological regions. As shown in Fig.~\ref{fig:counterfactual_reasoning}, the inferred counterfactual embeddings lie close to the corresponding real normal representations in the anatomical representation space, with cosine distances of only 0.003 and 0.005 in the two representative cases. This indicates that BaseAnatomy provides a sufficiently structured anatomical space for estimating the normal representation underlying a pathological patch. The resulting counterfactual representations therefore provide reliable references for calculating phenotype-specific increments in $\Delta$Phenotype.

\subsection{Continual Representation Learning Performance of Visual Phenotypes and t-SNE Visualization in Different Data Scale}

Table~\ref{tab:lesion_data_ratio} and Fig.~\ref{fig:different_scale_performance} examine the effect of lesion-positive training data on phenotype representation learning. As the training scale increases from 1\% to 100\%, phenotype accuracy rises from 25.80 to 43.11 for ReXGroundingCT VQA and from 15.55 to 52.65 for RRG. The corresponding accuracies on LIDC-IDRI increase from 54.55 to 77.69 and from 50.41 to 75.21. The representation projections show a parallel change: categories that overlap at smaller data scales form increasingly compact and distinguishable clusters as more lesion examples are included. In particular, atelectasis, consolidation, ground-glass opacity, and pulmonary nodules become more clearly separated. These results indicate that the encoder learns discriminative lesion representations and that additional lesion data improve both phenotype characterization and the organization of the learned representation space.

\subsection{Case Study}

Figure~\ref{fig:case_study} illustrates two representative cases in which the original backbone fails to identify the pulmonary abnormality. 
For both ground-glass opacity and atelectasis, the lesion patches are first located within the learned anatomical representation space, preserving their anatomical context. 
Their phenotype increments are then projected into the corresponding phenotype regions, providing discriminative evidence for pathological characterization. 
With these representations, our model correctly identifies both abnormalities and describes their bilateral lower-lung locations, whereas the original backbone reports no focal pulmonary abnormality. 
These examples show that $\Delta$Representation improves report generation by jointly preserving anatomical information and encoding phenotype-specific visual changes.

\section{Conclusion}

We introduced \textbf{$\Delta$Representation}, a visual representation learning framework for pathological phenotype interpretation in medical VLMs. \textbf{BaseAnatomy} structures normal anatomical representations with ideal spatial distributions, while \textbf{$\Delta$Phenotype} learns phenotype-specific representation increments through counterfactual reasoning. Experiments across two datasets and multiple medical VLM backbones show consistent improvements in lesion grounding and phenotype characterization, demonstrating the effectiveness of anatomy-referenced incremental representation learning for fine-grained medical image interpretation.

\subsection*{AI use statement}

In this work, we used generative AI tools for evaluating the report generation task. Specifically, GPT6-Astra was used as a structured parser to extract predicted lesion locations and phenotype labels from generated radiology reports. The extracted predictions were then compared with the reference annotations using the deterministic evaluation metrics described in Appendix~\ref{app:data}. GPT6-Astra was not used to assign performance scores directly. We did not use generative AI tools for generating synthetic datasets, developing theoretical models or conceptual frameworks, formulating mathematical claims, proposing or refining hypotheses, designing research methodology or experiments, implementing methods, assisting with translation, cleaning or reformatting datasets, or supporting qualitative and thematic data analysis. Proof-related uses are not applicable to this work. Additionally, we used generative AI tools for drafting parts of the research paper, improving readability, and suggesting the structure of the manuscript. We reviewed and revised all AI-assisted content, manually checked factual statements and citations against the original sources, and take full responsibility for the final content of this work, including text, claims, evaluations, and other artifacts produced with the aid of generative AI.

\subsection*{Ethics statement}

This work does not raise ethical concerns that require additional disclosure under the conference guidelines. It involves no new human-subject recruitment or collection of identifiable private information, and all datasets are used in accordance with their original access and usage requirements.

\subsection*{Reproducibility statement}

We provide an anonymous code repository at \url{https://anonymous.4open.science/r/deltarep-CF6D} to support reproduction of the methods and experiments presented in this work.

\bibliography{iclr2027_conference}
\bibliographystyle{iclr2027_conference}

\newpage
\appendix

\section{Benchmark and Evaluation}
\label{app:data}

\subsection{Benchmark Construction}
\begin{wrapfigure}{r}{0.52\linewidth}
\centering
\includegraphics[width=0.98\linewidth]{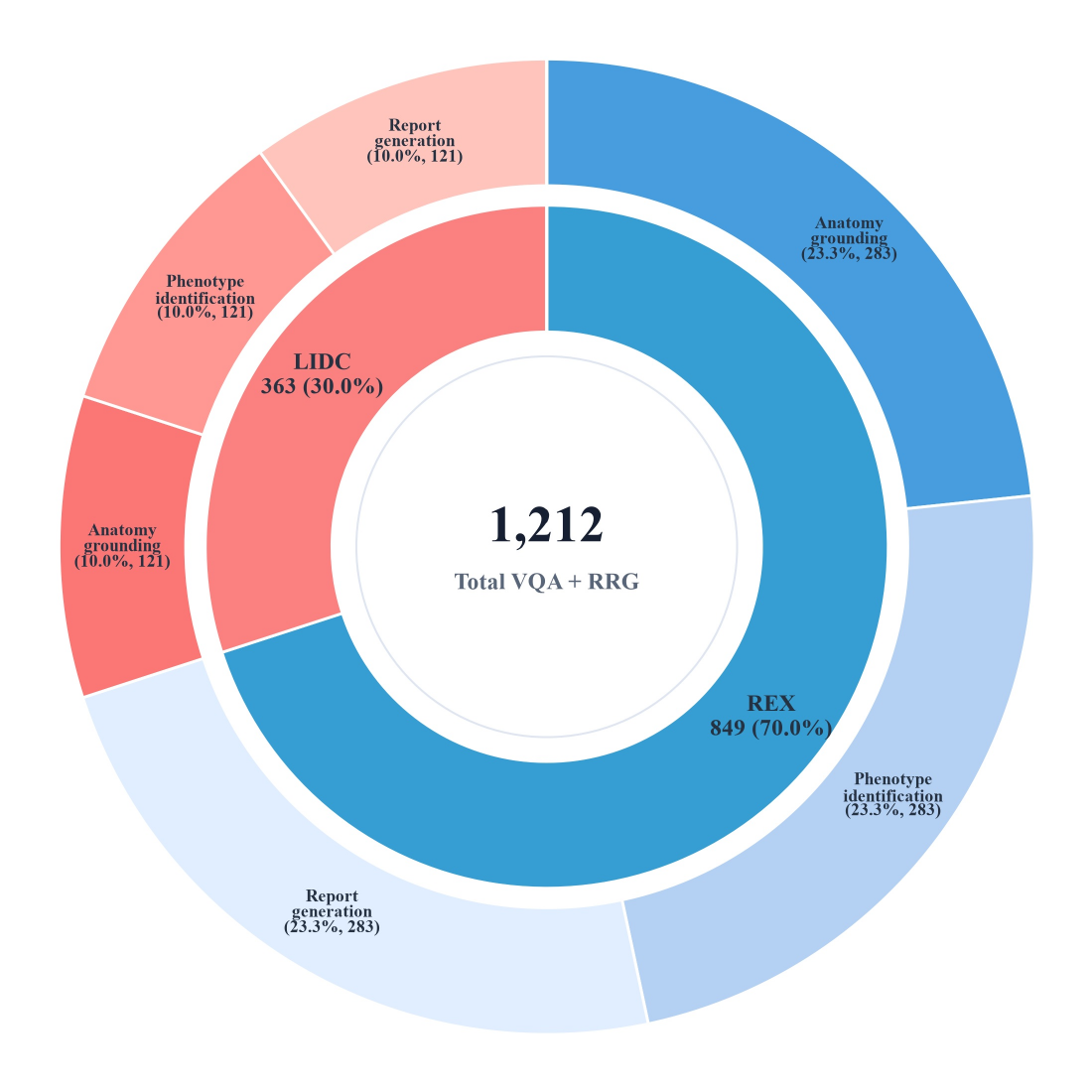}
\caption{Distribution of benchmark instances by dataset and task.}
\label{fig:benchmark_distribution}
\end{wrapfigure}

We construct a patient-disjoint benchmark from ReXGroundingCT and LIDC-IDRI. The unit of analysis is an unmarked axial chest CT image, rather than an entire CT volume. The curated benchmark contains 4,199 images: 3,233 from ReXGroundingCT and 966 from LIDC-IDRI. It is divided into 3,401 training, 394 development, and 404 locked test images, with no patient overlap between splits. The ReXGroundingCT training/development/test counts are 2,720/230/283; the corresponding LIDC-IDRI counts are 681/164/121. The ReXGroundingCT test set covers four target phenotypes: atelectasis (35 images), consolidation (43), ground-glass opacity (88), and pulmonary nodule (117). The LIDC-IDRI test set contains 71 nodule-positive images and 50 no-lesion controls.

We construct one lesion-grounding and one phenotype VQA question for each image. For grounding, the image is divided into a $4\times4$ grid, with rows A--D from top to bottom and columns 1--4 from left to right. For a positive image, the reference answer is the cell containing the largest area of the target lesion mask. Each question presents this cell and three distinct distractors, preferentially including nearby cells without the target lesion. For a no-lesion image, the four options include \textit{No lesion} and three grid cells. The phenotype question specifies the target cell for positive images and asks the model to identify the abnormality. Its positive answer belongs to the four ReXGroundingCT phenotype categories listed above; LIDC-IDRI positive answers are pulmonary nodules, while negative questions include a no-lesion option. Lesion annotations are used to construct reference answers offline and are not provided at test-time inference.

Each image also contributes one radiology report generation (RRG) prompt requesting a concise finding that states the primary abnormality and, when discernible, its laterality and craniocaudal region. For no-lesion images, the appropriate finding indicates the absence of a focal lesion. Unlike the VQA grounding task, RRG does not use or expose the \(4\times4\) grid.For no-lesion images, the appropriate finding indicates the absence of a focal lesion. Unlike the phenotype VQA question, the RRG prompt does not supply a target grid cell. The locked test set therefore comprises 808 VQA questions (566 ReXGroundingCT and 242 LIDC-IDRI) and 404 RRG instances (283 and 121, respectively), for 1,212 tasks in total (Fig.~\ref{fig:benchmark_distribution}).

\subsection{Evaluation Metrics}
We score generated answers separately for each dataset, using all $N_s$ test images from source $s$ ($N_{\mathrm{ReXGroundingCT}}=283$ and $N_{\mathrm{LIDC}}=121$) as the denominator of every metric. Let $g_i$ denote the reference grid cell or \textit{No lesion}, and let $d_i$ denote the reference phenotype or \textit{no lesion}. For VQA, \textit{Grounding Accuracy} and \textit{Phenotype Accuracy} are exact-match accuracies of the parsed generated answers:
\begin{equation}
\begin{aligned}
\operatorname{Acc}^{(s)}_{\mathrm{VQA\text{-}G}}
&=\frac{1}{N_s}\sum_{i\in\mathcal{T}_s}
\mathbf{1}\!\left[\hat g_i=g_i\right],\\
\operatorname{Acc}^{(s)}_{\mathrm{VQA\text{-}P}}
&=\frac{1}{N_s}\sum_{i\in\mathcal{T}_s}
\mathbf{1}\!\left[\hat d_i^{\mathrm{VQA}}=d_i\right],
\end{aligned}
\end{equation}
where $\mathcal{T}_s$ is the test set for source $s$.

For RRG, we use GPT6-Astra as a structured parser to extract the phenotype and
lesion-location attributes from each generated finding. Given a generated finding
$r_i$, the parser produces
$(\hat{d}^{\mathrm{RRG}}_i,\hat{\ell}_i)=P_{\mathrm{GPT}}(r_i)$,
where $\hat{d}^{\mathrm{RRG}}_i$ denotes the extracted phenotype and
$\hat{\ell}_i$ denotes the extracted lesion-location attributes, including
laterality and craniocaudal region. The extracted predictions are then compared
deterministically with the corresponding reference annotations. Missing,
ambiguous, or incorrect extracted attributes are treated as incorrect. For
no-lesion cases, an explicit statement indicating the absence of a focal lesion
is parsed as \textit{No lesion}. We compute Phenotype Accuracy and Grounding
Accuracy as

\begin{equation}
\begin{aligned}
\mathrm{Acc}^{(s)}_{\mathrm{RRG-P}}
&=
\frac{1}{N_s}
\sum_{i\in T_s}
\mathbf{1}
\left[
\hat{d}^{\mathrm{RRG}}_i=d_i
\right],\\
\mathrm{Acc}^{(s)}_{\mathrm{RRG-G}}
&=
\frac{1}{N_s}
\sum_{i\in T_s}
\mathbf{1}
\left[
\hat{\ell}_i=\ell_i
\right].
\end{aligned}
\label{eq:rrg_metrics}
\end{equation}

Here, $d_i$ and $\ell_i$ denote the reference phenotype and lesion-location
attributes, respectively. GPT6-Astra is used only to convert free-form findings
into structured predictions; metric scores are computed deterministically from
the extracted outputs.

\section{Disease-wise RRG\&VQA Results}
\label{app:diseasewise}

Table~\ref{tab:a2} decomposes the ReXGroundingCT results by phenotype for both VQA and RRG. The four disease subsets contain 35 atelectasis, 43 consolidation, 88 ground-glass opacity, and 117 pulmonary nodule cases. For each task, we report a composite accuracy obtained by pooling the Grounding and Phenotype decisions within the corresponding disease subset. Method rows report the absolute accuracy together with the gain over the corresponding native backbone.

\paragraph{Backbone-level consistency.}
The disease-wise breakdown shows that the aggregate improvement is not driven
by one phenotype or one language backbone. Adding $\Delta$Representation
improves nearly every disease-metric pair across RadFM, LLaVA-Med, Lingshu,
Hulu-Med, MedGemma, and MedGemma1.5, while the remaining cells are maintained.
The size of the gain differs across backbones because their native report
behavior is highly uneven: some models identify a disease but omit its
location, whereas others default to a narrow set of common findings. The
consistent direction of change indicates that the visual representation is
useful beyond the decoder on which it was originally developed.

\paragraph{MedGemma1.5 comparison.}
With the MedGemma1.5 text backbone fixed, this block compares different alignment strategies. Under the same MedGemma1.5 backbone, $\Delta$ Representation improves disease-wise performance across both VQA and RRG. The gains are observed across atelectasis, consolidation, ground-glass opacity, and pulmonary nodules, rather than being concentrated in a single phenotype. Compared with RadAlign, MCA-RG, and Reg2RG, $\Delta$Representation shows stronger VQA performance across the four phenotype subsets and remains consistently improved over the native backbone on RRG.

\paragraph{Residual difficulty by phenotype.}
The magnitude of improvement varies across phenotypes and backbones, indicating that representation learning does not affect all lesion appearances equally. Nevertheless, the improvements are distributed across all four phenotype subsets rather than being driven by a single disease category. This disease-wise analysis therefore complements the aggregate results in the main paper and shows that the overall gains persist across heterogeneous pathological appearances.

\definecolor{deltagain}{RGB}{20,93,132}
\newcommand{\gain}[1]{\,{\color{deltagain}\scriptsize$\uparrow$#1}}
\newcommand{\bestcell}[2]{\textbf{#1}\gain{#2}}

\begin{table}[!ht]
\centering
\setlength{\tabcolsep}{2.8pt}
\renewcommand{\arraystretch}{1.16}
\resizebox{\textwidth}{!}{%
\begin{tabular}{@{}ll*{8}{c}@{}}
\toprule
\multirow{2}{*}{Models} & \multirow{2}{*}{Scale}
& \multicolumn{2}{c}{Atelectasis Phenotype}
& \multicolumn{2}{c}{Consolidation Phenotype}
& \multicolumn{2}{c}{Ground-glass opacity Phenotype}
& \multicolumn{2}{c}{Pulmonary nodule Phenotype} \\
\cmidrule(lr){3-4}\cmidrule(lr){5-6}\cmidrule(lr){7-8}\cmidrule(l){9-10}
& & VQA Acc. & RRG Acc. & VQA Acc. & RRG Acc.
  & VQA Acc. & RRG Acc. & VQA Acc. & RRG Acc. \\
\midrule
RadFM & 14B
& 7.14 & 4.29 & 8.14 & 4.65 & 6.82 & 3.41 & 8.55 & 4.27 \\
\quad w/ $\Delta$Representation & $\sim$14B
& 44.29\gain{37.14} & 17.14\gain{12.86}
& 48.84\gain{40.70} & 18.60\gain{13.95}
& 36.93\gain{30.11} & 13.07\gain{9.66}
& 49.15\gain{40.60} & 14.10\gain{9.83} \\
\addlinespace
LLaVA-Med & 7B
& 28.57 & 24.29 & 29.07 & 29.07 & 26.70 & 18.75 & 27.78 & 23.93 \\
\quad w/ $\Delta$Representation & $\sim$7B
& 55.71\gain{27.14} & 45.71\gain{21.43}
& 72.09\gain{43.02} & 45.35\gain{16.28}
& 54.55\gain{27.84} & 30.68\gain{11.93}
& 59.40\gain{31.62} & 39.74\gain{15.81} \\
\addlinespace
Lingshu & 7B
& 51.43 & 1.43 & 56.98 & 2.33 & 40.34 & 2.27 & 45.30 & 2.56 \\
\quad w/ $\Delta$Representation & $\sim$7B
& 64.29\gain{12.86} & 15.71\gain{14.29}
& 69.77\gain{12.79} & 19.77\gain{17.44}
& 56.82\gain{16.48} & 14.77\gain{12.50}
& 65.81\gain{20.51} & 16.24\gain{13.68} \\
\addlinespace
Hulu-Med & 4B
& 45.71 & 4.29 & 55.81 & 5.81 & 45.45 & 5.11 & 48.29 & 5.98 \\
\quad w/ $\Delta$Representation & $\sim$4B
& 57.14\gain{11.43} & 15.71\gain{11.43}
& 72.09\gain{16.28} & 18.60\gain{12.79}
& 55.68\gain{10.23} & 15.34\gain{10.23}
& 67.09\gain{18.80} & 14.96\gain{8.97} \\
\addlinespace
MedGemma & 4B
& 41.43 & 1.43 & 45.35 & 2.33 & 37.50 & 1.14 & 42.31 & 2.56 \\
\quad w/ $\Delta$Representation & $\sim$4B
& 55.71\gain{14.29} & 15.71\gain{14.29}
& 62.79\gain{17.44} & 20.93\gain{18.60}
& 47.16\gain{9.66} & 14.20\gain{13.07}
& 63.68\gain{21.37} & 13.68\gain{11.11} \\
\midrule
MedGemma1.5 & 4B
& 30.00 & 2.86 & 37.21 & 3.49 & 25.57 & 2.84 & 29.91 & 2.99 \\
\quad w/ RadAlign & $\sim$4B
& 31.43\gain{1.43} & 31.43\gain{28.57}
& 38.37\gain{1.16} & 40.70\gain{37.21}
& 26.14\gain{0.57} & 27.27\gain{24.43}
& 30.77\gain{0.85} & 29.49\gain{26.50} \\
\quad w/ MCA-RG & $\sim$4B
& 42.86\gain{12.86} & 34.29\gain{31.43}
& 51.16\gain{13.95} & 48.84\gain{45.35}
& 36.93\gain{11.36} & 28.98\gain{26.14}
& 45.73\gain{15.81} & 43.59\gain{40.60} \\
\quad w/ Reg2RG & $\sim$4B
& 41.43\gain{11.43} & 35.71\gain{32.86}
& 52.33\gain{15.12} & 46.51\gain{43.02}
& 36.36\gain{10.80} & 28.98\gain{26.14}
& 45.73\gain{15.81} & 44.02\gain{41.03} \\
\quad w/ $\Delta$Representation & $\sim$4B
& \bestcell{67.14}{37.14} & \bestcell{38.57}{35.71}
& \bestcell{61.63}{24.42} & \bestcell{52.33}{48.84}
& \bestcell{55.11}{29.55} & \bestcell{31.25}{28.41}
& \bestcell{64.53}{34.62} & \bestcell{44.87}{41.88} \\
\bottomrule
\end{tabular}%
}
\caption{\textbf{Disease-wise VQA and RRG performance on ReXGroundingCT.}
Values are composite accuracies (\%), obtained by pooling the Grounding and
Phenotype decisions within each task. Gains are absolute percentage points
over the native backbone.}
\label{tab:a2}
\end{table}

\section{High-Dimensional Representation Diagnostics}
\label{app:representation_diagnostics}

Table \ref{tab:representation-quantitative} provides quantitative diagnostics of the learned representation space in its original high-dimensional form, complementing the qualitative projections in the main paper. We evaluate three properties that correspond to the central objectives of $\Delta$Representation: the geometric stability of BaseAnatomy, the separability of phenotype increments, and the quality of the estimated counterfactual normal references. We further examine how these representations change under spatial pooling.

\paragraph{Anatomical representation stability.}
BaseAnatomy shows consistently high agreement under rotation, scaling, and translation, with Pearson correlations of 0.9742, 0.9751, and 0.9750, respectively. These results indicate that the learned anatomical representations preserve their relative organization under moderate spatial transformations rather than depending on a specific image configuration. This quantitatively supports the cross-transformation consistency observed in the anatomical representation analysis.

\paragraph{Phenotype organization and counterfactual normal references.}
The learned phenotype increments achieve a cosine silhouette score of 0.3607 and a nearest-prototype accuracy of 75.46\% on the ReX four-phenotype evaluation set. These measurements show that phenotype-specific increments exhibit meaningful separation in the original representation space, consistent with the progressively clearer phenotype organization visualized in Fig.~5. For counterfactual normal estimation, the normal-ring representations remain highly compact, with a compactness score of 0.9882. The mean and median cosine distances between abnormal representations and their estimated normal references are 0.0233 and 0.0210, respectively. Together, these results indicate that BaseAnatomy provides a locally coherent anatomical space from which stable normal references can be estimated for pathological patches.

\paragraph{Sensitivity to spatial pooling.}
We additionally evaluate representation behavior before and after exact $4\times4$ spatial pooling. Fixed 5-NN accuracy increases from 60.98\% to 64.23\%, while macro accuracy increases from 58.98\% to 60.83\%. In contrast, the cosine silhouette and own-centroid cosine decrease modestly after pooling. This pattern suggests that pooling preserves, and in the local classification metrics slightly improves, discriminative neighborhood structure while introducing a small reduction in within-class geometric compactness. Therefore, the pooled representations are not strictly invariant to spatial aggregation, but retain the anatomical and phenotype-related organization required by the downstream visual-language interface.

Overall, the high-dimensional measurements support the qualitative observations in the main paper from three complementary perspectives. BaseAnatomy maintains stable anatomical geometry under spatial transformations, $\Delta$Phenotype produces separable phenotype-specific increments, and the resulting representations remain informative after the spatial pooling used for downstream VLM alignment.

\begin{table}[t]
\centering
\small
\setlength{\tabcolsep}{4pt}
\renewcommand{\arraystretch}{1.08}

\begin{tabular*}{\linewidth}{@{\extracolsep{\fill}}>{\raggedright\arraybackslash}p{0.18\linewidth}>{\raggedright\arraybackslash}p{0.30\linewidth}>{\raggedright\arraybackslash}p{0.30\linewidth}r@{}}
\multicolumn{4}{@{}l}{\textbf{(a) Absolute representation quality}} \\
\toprule
Component & Evaluation cohort & Metric & Score \\
\midrule
BaseAnatomy
& ReX supervised pool
& Rotation Pearson $r$ $\uparrow$
& 0.9742 \\
& & Scaling Pearson $r$ $\uparrow$ & 0.9751 \\
& & Translation Pearson $r$ $\uparrow$ & 0.9750 \\
\addlinespace[2pt]
$\Delta$Phenotypes
& ReX four-phenotype set, $n=379$
& Cosine silhouette $\uparrow$
& 0.3607 \\
& & Nearest-prototype accuracy $\uparrow$ & 75.46\% \\
\addlinespace[2pt]
Normal reference
& LIDC validation, 232 cases / 1,120 lesion patches
& Normal-ring compactness $\uparrow$
& 0.9882 \\
& & Abnormal-to-normal cosine distance (mean) & 0.0233 \\
& & Abnormal-to-normal cosine distance (median) & 0.0210 \\
\bottomrule
\end{tabular*}

\vspace{5pt}

\begin{tabular*}{\linewidth}{@{\extracolsep{\fill}}>{\raggedright\arraybackslash}p{0.28\linewidth}>{\raggedright\arraybackslash}p{0.26\linewidth}rrr@{}}
\multicolumn{5}{@{}l}{\textbf{(b) Scale sensitivity}} \\
\toprule
Metric & Evaluation cohort & 1$\times$ & 4$\times$ & $\Delta$ \\
\midrule
Fixed 5-NN accuracy $\uparrow$
& ReX validation, $n=123$
& 60.98\% & \textbf{64.23\%} & +3.25 pp \\
Macro accuracy $\uparrow$
& ReX validation, $n=123$
& 58.98\% & \textbf{60.83\%} & +1.85 pp \\
Cosine silhouette $\uparrow$
& ReX public test, $n=55$
& \textbf{0.5397} & 0.5157 & $-0.0240$ \\
Own-centroid cosine $\uparrow$
& ReX public test, $n=55$
& \textbf{0.9424} & 0.9203 & $-0.0221$ \\
\bottomrule
\end{tabular*}
\caption{High-dimensional representation diagnostics. Panel (a) reports
absolute representation-quality measurements. Panel (b) isolates the effect
of exact $4\times4$ spatial pooling. Here, $\Delta=4\times-1\times$ and pp
denotes percentage points.}
\label{tab:representation-quantitative}
\end{table}

\section{Implementation Details}
\label{app:implementation}

\paragraph{Backbone initialization and input.}
For each evaluated medical VLM, the anatomical encoder $E_A$ is initialized from
its native vision tower, and $\Delta$Representation is trained separately for each
backbone. After BaseAnatomy training, $E_A$ is frozen and the phenotype encoder
$E_P$ is initialized from the same BaseAnatomy checkpoint, following
Sec.~3.3. The numerical settings below describe the MedGemma~1.5--4B
configuration used for the detailed representation analyses; architecture-dependent
dimensions follow the corresponding native vision tower for the other VLMs.
For MedGemma~1.5--4B, each axial CT slice is resized to $896\times896$,
normalized to $[-1,1]$, and divided into a $64\times64$ grid with patch size 14.
The current slice index and total slice count are used to encode the relative
craniocaudal position, and patch embeddings are $\ell_2$-normalized.

\paragraph{BaseAnatomy training.}
BaseAnatomy follows the geometry-supervised learning objective described in
Sec.~3.2. We set $\alpha=0.65$ in Eq.~(1) and sample anatomical classes in a
balanced manner. Rotation and scale transformations are applied with matched
spatial correspondence, using rotations in $[-10^{\circ},10^{\circ}]$ and
isotropic scaling in $[0.92,1.08]$. Anatomical annotations define normal tissue
regions, while lesion masks exclude pathological patches from anatomical
supervision. No additional representation-learning objectives beyond those
described in Sec.~3.2 are introduced.

\paragraph{$\Delta$Phenotype training.}
After BaseAnatomy converges, $E_A$ remains frozen and $E_P$ is optimized to
encode phenotype-specific pathological changes as described in Sec.~3.3. For
each pathological patch, its counterfactual normal reference is estimated from
8--48 anatomically matched, mask-negative neighbors within grid radii 4--10
using normalized inverse-distance weighting. Phenotype-specific consistency and
separation, lesion localization, anatomical preservation, and no-lesion
suppression follow the objectives described in Sec.~3.3. No additional
phenotype-learning objectives are introduced.

\paragraph{Optimization.}
Both representation-learning stages use AdamW in BF16 with weight decay 0.01,
cosine learning-rate decay, and gradient clipping at 1.0. Checkpoints are
selected only on the patient-disjoint development split.

\paragraph{Vision--language alignment and inference.}
After representation learning, only $E_P$ is retained as the VLM vision encoder.
For downstream VQA and RRG, the trained vision encoder and text decoder are
frozen and only the corresponding multimodal projector is optimized. For the
MedGemma~1.5--4B configuration, each $4\times4$ block of the $64\times64$
visual grid is average-pooled, yielding 256 visual tokens. Alignment uses batch
size two, gradient accumulation over two steps, a learning rate of
$1.2\times10^{-5}$, and two epochs. At inference, the model receives only the
unmarked CT slice, slice-position information, and text prompt. Anatomical
masks, lesion masks, and counterfactual normal-reference estimation are used
only during representation learning and are not provided at inference time.

\paragraph{Compute.}
Representation learning uses one or two NVIDIA RTX PRO 6000 Blackwell Server
Edition GPUs, while alignment and evaluation additionally use an NVIDIA RTX
A6000 with 48~GB memory.

\end{document}

%% file: math_commands.tex
\usepackage{amsmath,amsfonts,bm}

\def\eqref#1{equation~\ref{#1}}

\def\1{\bm{1}}

\DeclareMathAlphabet{\mathsfit}{\encodingdefault}{\sfdefault}{m}{sl}
\SetMathAlphabet{\mathsfit}{bold}{\encodingdefault}{\sfdefault}{bx}{n}

